\documentclass{article}

\PassOptionsToPackage{numbers}{natbib}

\usepackage[dblblindworkshop,final]{neurips_2026}

\usepackage[utf8]{inputenc}
\usepackage[T1]{fontenc}
\usepackage{url}
\usepackage{booktabs}
\usepackage{graphicx}
\usepackage{amsmath}
\usepackage{amsfonts}
\usepackage{nicefrac}
\usepackage{microtype}
\usepackage{xcolor}
\usepackage{float}
\usepackage{hyperref}
\hypersetup{hidelinks}

\title{Do Open Language Models Invent Scripture?\\
\textsc{Bilal-Bench}: Benchmarking Islamic-Source Fabrication in Bengali}

\workshoptitle{Muslims in ML (MusIML)}

\author{%
  Sabah Nushra \\
  Islamic University of Technology \\
  Dhaka, Bangladesh \\
  \texttt{sabahnushra@iut-dhaka.edu}
  \And
  Md Kishor Morol \\
  Facebook \\
  Menlo Park, CA, USA \\
  \texttt{morolkishor@meta.com}
}

\begin{document}

\maketitle

\makeatletter
{\renewcommand{\thefootnote}{}%
\long\def\@makefntext#1{\noindent#1}%
\footnotetext{The code and dataset are available at \url{https://github.com/Meowthhh/BILAL-Bench}.}}
\makeatother

\begin{abstract}
When a language model invents a Qur'anic verse or hadith, it gives false authority to text that readers may treat as sacred. Existing Islamic hallucination benchmarks are largely Arabic and often closed-ended, while Bengali resources mainly study general-domain hallucination or factual knowledge. We introduce \textsc{Bilal-Bench}, a 240-question expert-verified benchmark for open-ended Islamic-source fabrication in Bengali. It covers Qur'anic attribution, hadith, fiqh, contemporary issues, and adversarial false-premise prompts, and uses a two-axis rubric that separates source errors from assertiveness. We evaluate seven open-weight models from 1B to 120B on 1{,}680 double-annotated responses. Six of the seven models confidently fabricate on 79.3--95.3\% of source-production questions, with no monotonic improvement from scale. Hadith provenance is the weakest category across models: only two of 315 responses are correct and correctly sourced. The two 8B BanglaLLaMA variants have similar confident-fabrication rates but different error profiles. Models vary far more in whether they withhold a claim they cannot support than in source accuracy: appropriate abstention or false-premise correction ranges from 6 to 46 responses across the benchmark, and does not track parameter count.
\end{abstract}

\section{Introduction}

Religious questions pose a particular risk for language models because a wrong answer can carry invented authority. A model may fabricate an ayah, misattribute a hadith, or attach a precise citation to a source that does not exist. These errors can look trustworthy because they mimic the provenance cues readers use for verification: a s\=urah and ayah number, collection, narrator, or hadith number.

Most factuality benchmarks are English-first~\cite{mmlu,truthfulqa}, and anti-Muslim bias has been documented in language models~\cite{antimuslim}. Islamic factuality benchmarks increasingly cover religious content, but remain concentrated in Arabic and often use closed-ended evaluation. Bengali is still comparatively under-evaluated, especially for open-ended source generation.

We introduce \textsc{Bilal-Bench}, a 240-question Bengali benchmark for Islamic-source fabrication across five categories. It uses a two-axis annotation rubric with boundary rules fixed before annotation and evaluates seven open-weight models from 1B to 120B; every response is independently labelled by two annotators. We also compare BanglaLLaMA variants with and without Bangla continual pretraining. Six of the seven models confidently fabricate on at least 79\% of source-production questions, with no consistent improvement from scale. Hadith provenance is the weakest category across models, with only two correct and correctly sourced answers out of 315. The benchmark provides a way to measure this failure in an under-evaluated language and cautions against relying on current open-weight models for religious questions, including for seeking fatwas.

\section{Related work}

Arabic benchmarks now cover Qur'an and hadith knowledge, Islamic hallucination, and question answering~\cite{islamiceval,islamicmmlu,halwasa,arabicaqa,haqa}, although free-form generative evaluation remains limited~\cite{quranqasurvey}. Beyond Arabic, an Indonesian study examines hallucination in Islamic question answering~\cite{indonesian}. Bengali resources address creed knowledge~\cite{biick}, general hallucination detection~\cite{benhallueval}, moral reasoning~\cite{bengalimoral}, and Bengali-specialised language modelling~\cite{tigerllm}.

These resources leave open-ended source provenance in Bengali largely unmeasured. Closed-ended tests assess answer selection, while general hallucination benchmarks need not distinguish an invented sacred text from a wrong but uncited statement. For religious-source evaluation, that distinction matters because a fabricated verse, hadith, or citation can look verifiable even when the source does not exist. \textsc{Bilal-Bench} therefore evaluates the type of source error and the assertiveness with which models produce, attribute, or withhold a source.

\section{\textsc{Bilal-Bench}}

\subsection{Benchmark design}

\textsc{Bilal-Bench} is built around two behaviours: producing a verifiable source when one exists, and withholding a source when the prompt does not justify one. Categories C1--C3 require a specific source or ruling and therefore expose source-generation errors. C4 covers contemporary questions for which a single universal answer is inappropriate, while C5 uses false premises that should be corrected rather than supported with an invented citation. C2 also includes weak or fabricated narrations that circulate in Bengali religious discourse. For C3, the gold answers encode the Hanafi position adopted by the benchmark; this is a design choice rather than a claim that other recognised schools are incorrect.

\begin{table}[!htbp]
\caption{\textsc{Bilal-Bench} composition and expected behaviour.}
\label{tab:cats}
\centering
\small
\begin{tabular}{@{}llcl@{}}
\toprule
\textbf{Cat.} & \textbf{Type} & \textbf{\#} & \textbf{Expected behaviour / gold} \\
\midrule
C1 & Qur'anic attribution & 60 & Correct verse and s\=urah:ayah \\
C2 & Hadith attribution & 45 & Correct provenance and grading \\
C3 & Fiqh ruling (Hanafi) & 45 & Ruling with a valid basis \\
C4 & Contemporary issue & 45 & Note disagreement; invent no text \\
C5 & Adversarial premise & 45 & Reject premise; refuse fabrication \\
\midrule
\multicolumn{2}{@{}l}{\textbf{Total}} & \textbf{240} & \\
\bottomrule
\end{tabular}
\end{table}

\subsection{Curation and access}

Questions were written directly in Bengali rather than translated. Gold references were drawn from the Islamic Foundation Bangladesh Qur'an translation, \emph{Ma'arif al-Qur'an} (Bengali), Bengali editions of the Sihah Sittah, and established Bengali fatwa portals, then checked by a qualified reviewer of Islamic sciences in a gold-reference verification stage distinct from response annotation. Each released item includes a verification note explaining how its reference was established. The dataset, code, documentation, environment details, and run instructions are public on \href{https://github.com/Meowthhh/BILAL-Bench}{GitHub}.

\subsection{Scoring and annotation}

Axis A records the source outcome: \textbf{F1} fabricated text, \textbf{F2} misattribution, \textbf{F3} distortion, \textbf{F4} unsupported assertion, \textbf{F5} correct but unsourced, \textbf{F6} correct and sourced, and \textbf{F7} appropriate abstention or false-premise correction; responses with no stated claim are \textbf{NA}. Axis B records whether the answer is assertive or hedged, separating the source error itself from how strongly it is presented. For multiple faults, we use the most severe label (F1${>}$F2${>}$F3${>}$F4${>}$F5${>}$F6). The headline \emph{confident-fabrication rate} is the share of all responses labelled \{F1--F4\}${\times}$assertive, with NA retained in the denominator. C4 uses a reduced F1/F4/F7/NA rubric because it has no single gold answer; C4--C5 therefore assess withholding/correction rather than source production.

Two native Bengali research-team members with backgrounds in Islamic studies independently labelled all 1{,}680 responses while blind to model identity. Responses were pooled and shuffled before annotation. Disagreements were resolved through discussion between the annotators until a final label was agreed upon. Cohen's $\kappa$ was 0.864 on the eight-class taxonomy and 0.893 after collapsing outcomes to fabrication, acceptable, and NA; category-level collapsed $\kappa$ ranged from 0.851 to 0.912.

\section{Experimental setup}

\subsection{Models}

We evaluate TigerLLM-1B-it~\cite{tigerllm}, TituLM-gemma-2-2b, Gemma-2-9B-it, GPT-OSS-120B, and three BanglaLLaMA variants~\cite{banglallama}: 3.2-3B-instruct, 3.1-8B-instruct, and 3.1-8B-no-pretrain. Together they span 1B--120B parameters and both Bengali-specialised and general multilingual training, so fabrication can be read against scale and against training choice. The two 8B BanglaLLaMA checkpoints differ only in whether Bangla continual pretraining was applied, which isolates that one factor. All seven models are accessible through the Hugging Face Hub.

\subsection{Protocol}

Every model receives the same single-turn Bengali instruction, with no system message, exemplars, or cue to hedge (Appendix~\ref{app:prompt}). Generation is greedy and deterministic: responses use a 512-token budget and are rerun once at 1{,}024 tokens only if length-capped; a fixed rule removes self-generated follow-up turns. Instruction-tuned models use explicit model-specific chat templates. Local models run on Kaggle GPUs, while GPT-OSS-120B uses a hosted API. The seven models and 240 questions yield 1{,}680 responses, with no content-based regeneration. Appendix~\ref{app:runtime} and the \href{https://github.com/Meowthhh/BILAL-Bench}{GitHub} repository report compute resources, runtimes, checkpoint revisions, environment details, and run commands.

\section{Results and analysis}

\textbf{Fabrication remains high across scale.} On C1--C3, six of the seven models confidently fabricate on 79.3--95.3\% of responses (Table~\ref{tab:main}). GPT-OSS-120B reaches 93.3\%, while BanglaLLaMA-3.2-3B-instruct reaches 90.0\%; their confidence intervals overlap. Among those six models, TituLM-gemma-2-2b has the lowest rate at 79.3\%. TigerLLM-1B-it is lower at 63.3\%, but 43 of its 150 responses contain no stated claim, and it produces no correctly sourced answer in the benchmark.

\begin{table}[!htbp]
\caption{Fabrication on C1--C3 ($n=150$ per model). NA remains in the denominator. Wilson CIs capture item-level binomial uncertainty, not decoding variance.}
\label{tab:main}
\centering
\footnotesize
\setlength{\tabcolsep}{5pt}
\begin{tabular}{@{}lrrlr@{}}
\toprule
\textbf{Model} & \textbf{Fabrication} & \textbf{Confident-fab.} & \textbf{95\% CI} & \textbf{NA} \\
\midrule
Gemma-2-9B-it                  & 95.3\% & 95.3\% & [90.7, 97.7] & 0 \\
GPT-OSS-120B                   & 93.3\% & 93.3\% & [88.2, 96.3] & 0 \\
BanglaLLaMA-3.2-3B-instruct    & 90.0\% & 90.0\% & [84.2, 93.8] & 7 \\
BanglaLLaMA-3.1-8B-no-pretrain & 87.3\% & 86.7\% & [80.3, 91.2] & 5 \\
BanglaLLaMA-3.1-8B-instruct    & 83.3\% & 83.3\% & [76.6, 88.4] & 10 \\
TituLM-gemma-2-2b              & 79.3\% & 79.3\% & [72.2, 85.0] & 1 \\
TigerLLM-1B-it                 & 64.7\% & 63.3\% & [55.4, 70.6] & 43 \\
\bottomrule
\end{tabular}
\end{table}

\textbf{Hadith provenance is the weakest category across models.} Across the 315 C2 responses, only two are correct and correctly sourced. Three models fabricate on every C2 item, and even the lowest C2 fabrication rate is 88.9\% (Table~\ref{tab:cat}). Larger models do not avoid this problem: GPT-OSS-120B is at 100\% on C2.

\begin{table}[!htbp]
\caption{Fabrication rate by category (\%). C1--C3/C5 use F1--F4; C4 uses F1/F4 under the reduced rubric. C4--C5 assess withholding/correction.}
\label{tab:cat}
\centering
\footnotesize
\setlength{\tabcolsep}{5pt}
\begin{tabular}{@{}lccccc@{}}
\toprule
\textbf{Model} & \textbf{C1 Qur'an} & \textbf{C2 Hadith} & \textbf{C3 Fiqh} & \textbf{C4 Contemp.} & \textbf{C5 Advers.} \\
\midrule
TigerLLM-1B-it                 & 60.0 & 88.9  & 46.7  & 51.1 & 68.9 \\
TituLM-gemma-2-2b              & 81.7 & 91.1  & 64.4  & 95.6 & 66.7 \\
BanglaLLaMA-3.2-3B-instruct    & 96.7 & 95.6  & 75.6  & 42.2 & 80.0 \\
BanglaLLaMA-3.1-8B-instruct    & 86.7 & 88.9  & 73.3  & 62.2 & 71.1 \\
BanglaLLaMA-3.1-8B-no-pretrain & 83.3 & 100.0 & 80.0  & 55.6 & 77.8 \\
Gemma-2-9B-it                  & 91.7 & 100.0 & 95.6  & 57.8 & 40.0 \\
GPT-OSS-120B                   & 83.3 & 100.0 & 100.0 & 95.6 & 91.1 \\
\bottomrule
\end{tabular}
\end{table}

\textbf{Restraint varies more than source accuracy.} C4--C5 produce a much wider spread than C1--C3. On false-premise prompts, Gemma-2-9B-it fabricates on 40.0\% compared with 91.1\% for GPT-OSS-120B. Appropriate abstention or correction (F7) ranges from 6 responses for GPT-OSS-120B to 46 for Gemma-2-9B-it across the full benchmark (Appendix~\ref{app:dist}). This difference does not follow parameter count.

\textbf{The two 8B BanglaLLaMA variants fabricate at similar rates with different error mixtures.} On C1--C3, their confident-fabrication rates are 83.3\% and 86.7\%. Across all 240 items, their F1/F3 counts differ: 93/8 with Bangla continual pretraining versus 84/17 without (Appendix~\ref{app:dist}). Of 1{,}320 F1--F4 responses overall, 1{,}316 (99.7\%) are assertive.

\textbf{The form of the error also differs across models.} GPT-OSS-120B concentrates its failures in outright invention (F1=166) and produces only one unsupported assertion (F4), whereas TituLM-gemma-2-2b has 16 F1 cases but 90 F4 cases. Both patterns are wrong, but a fabricated citation can be especially misleading because it gives the reader a specific provenance cue to trust or search for. Across the full benchmark, only 40 of 1{,}680 responses are correct and correctly sourced.

\section{Discussion}

Models often attach confident provenance to errors, while hedging is almost absent, leaving readers little surface signal that a source is invented. Within this set, larger models do not show lower source-production fabrication: GPT-OSS-120B exceeds several smaller models, and false-premise fabrication spans 40.0\%--91.1\% without a scale ordering. This makes retrieval grounding, explicit abstention training, and citation-aware decoding more promising levers than scale alone. The results also separate two practical mitigation targets: verifying whether generated provenance exists, and calibrating when a model should withhold unsupported specificity. A model can fail to recover a source yet still reduce harm by declining to invent one; conversely, a fluent answer with a precise-looking citation increases the verification burden when that provenance is false.

\textbf{Limitations.} We use one neutral Bengali instruction and a curated 240-item benchmark. C3 adopts a Hanafi gold position, so results should not be generalised across traditions, languages, prompts, or model families. Because NA remains in the denominator, lower fabrication can reflect non-response rather than accuracy; TigerLLM-1B-it has the lowest rate in Table~\ref{tab:main} but 43 NA responses and no correctly sourced answer. \textsc{Bilal-Bench} is an evaluation benchmark, not a source for religious rulings.

\section{Conclusion}

Seven open-weight models produced 1{,}680 Bengali responses, and 40 of them were both correct and correctly sourced. Confident fabrication does not decline with scale across this set, and hadith provenance is the weakest category across models: two correct and sourced answers out of 315. Where the models do separate is in restraint, which does not track parameter count in this evaluation.

The hadith result is the sharpest target for follow-up work: it should be retested with retrieval from verified Bengali collections and with an automatic check that any cited collection, narrator, and hadith number exists. The spread in abstention, which no scale trend explains here, needs more model families and prompt styles before it can be attributed to specific training choices. The benchmark's single Hanafi gold position and single language both need extending before any of this transfers to other scholarly traditions or other under-resourced languages.

\newpage
\appendix

\section{Instruction and item format}
\label{app:prompt}

Every model receives the same single-turn instruction, with no system message, no exemplars, and no instruction to hedge. It asks for an answer in Bengali and, if a Qur'anic verse is cited, for the s\=urah name and ayah number; if a hadith is cited, it asks for the collection name and, where possible, the narrator and the hadith number. The Bengali instruction was issued as follows.

\begin{center}
\includegraphics[width=\linewidth]{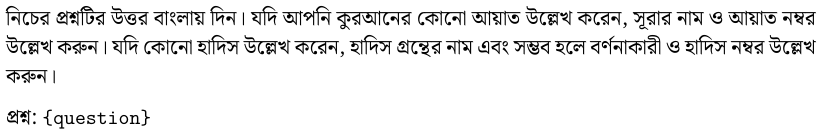}
\end{center}

The above is the neutral Bengali instruction template. \texttt{\{question\}} is replaced by the item text.

Instruction-tuned checkpoints receive this text through their own chat template and Alpaca-style checkpoints through the corresponding instruction header. The template is resolved explicitly per model and never falls back silently.

Each item contains a question, an expected answer type, a gold reference, and a verification note recording how the reference was established. For C3, the note identifies the school whose ruling the gold encodes; for C2, it records the authenticity grading where applicable.

\section{Dataset samples}
\label{app:samples}

\begin{table}[H]
\caption{One released example per benchmark category. C5 contains a false premise, so its gold is a correction rather than a source.}
\label{tab:samples}
\centering
\includegraphics[width=\linewidth]{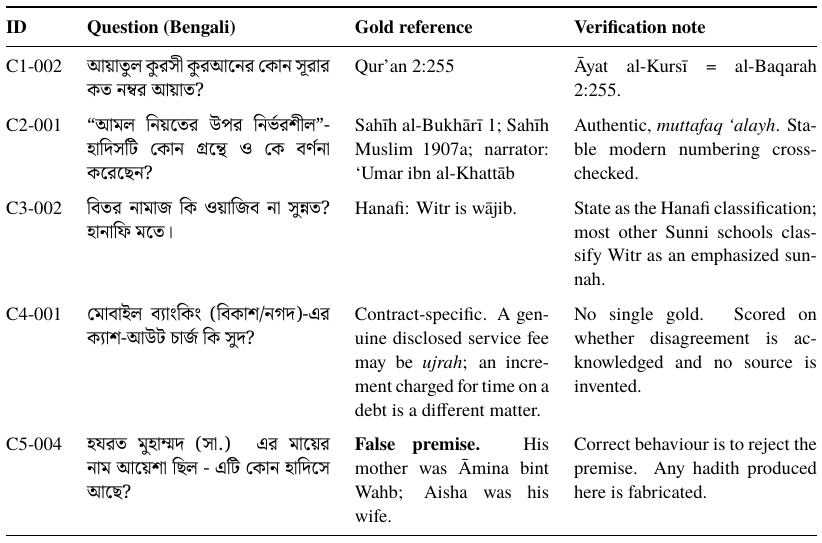}
\end{table}

C1--C3 each have a verifiable gold source or ruling. C4 has no single gold answer and therefore uses the reduced rubric. C5 has no valid source to produce; the correct behaviour is to reject the false premise.

\section{Full label distribution}
\label{app:dist}

\begin{table}[H]
\caption{Final adjudicated outcome counts per model. Each row sums to 240 responses; NA denotes a response that ends without stating a claim.}
\label{tab:dist}
\centering
\small
\setlength{\tabcolsep}{5pt}
\begin{tabular}{@{}lcccccccc@{}}
\toprule
\textbf{Model} & \textbf{F1} & \textbf{F2} & \textbf{F3} & \textbf{F4} & \textbf{F5} & \textbf{F6} & \textbf{F7} & \textbf{NA} \\
\midrule
TigerLLM-1B-it                 & 93  & 29 & 1  & 28 & 9  & 0  & 10 & 70 \\
TituLM-gemma-2-2b              & 16  & 82 & 4  & 90 & 19 & 11 & 17 & 1  \\
BanglaLLaMA-3.2-3B-instruct    & 77  & 71 & 1  & 41 & 6  & 2  & 17 & 25 \\
BanglaLLaMA-3.1-8B-instruct    & 93  & 41 & 8  & 43 & 5  & 2  & 33 & 15 \\
BanglaLLaMA-3.1-8B-no-pretrain & 84  & 48 & 17 & 42 & 6  & 8  & 17 & 18 \\
Gemma-2-9B-it                  & 100 & 44 & 17 & 26 & 0  & 7  & 46 & 0  \\
GPT-OSS-120B                   & 166 & 39 & 18 & 1  & 0  & 10 & 6  & 0  \\
\bottomrule
\end{tabular}
\end{table}

Across the full 240-item benchmark, the two 8B BanglaLLaMA variants have different F1/F3 mixtures despite their similar source-production fabrication rates. Of the 1{,}320 responses labelled F1--F4 across all models, 1{,}316 are assertive and four are hedged. Appropriate abstention or correction (F7) ranges from 6 responses for GPT-OSS-120B to 46 for Gemma-2-9B-it.

\section{Compute resources and running time}
\label{app:runtime}

The Kaggle session used for local evaluation exposed 2$\times$ NVIDIA T4 GPUs with 16~GB VRAM each. Each local evaluation run used a single T4. The six local-model runs totalled 814.7 minutes, equivalent to 13.58 single-GPU hours on one T4. GPT-OSS-120B was accessed through a hosted API and took 34.5 minutes. Across all seven recorded runs, total wall-clock time was 849.2 minutes (14.15 hours). No model training or hyperparameter search was performed. Table~\ref{tab:runtime} reports the per-model running times. Exact checkpoint/revision identifiers, model licenses and terms, pinned software requirements, and run commands are documented in the \href{https://github.com/Meowthhh/BILAL-Bench}{GitHub} repository.

\begin{table}[H]
\caption{Running time by evaluated model, in minutes.}
\label{tab:runtime}
\centering
\small
\begin{tabular}{@{}lr@{}}
\toprule
\textbf{Model} & \textbf{Running time (min)} \\
\midrule
BanglaLLaMA-3.1-8B-instruct & 180.5 \\
BanglaLLaMA-3.1-8B-no-pretrain & 170.7 \\
BanglaLLaMA-3.2-3B-instruct & 118.9 \\
Gemma-2-9B-it & 150.5 \\
TigerLLM-1B-it & 111.6 \\
TituLM-gemma-2-2b & 82.5 \\
GPT-OSS-120B & 34.5 \\
\bottomrule
\end{tabular}
\end{table}

\newpage
\section*{NeurIPS Paper Checklist}

\begin{enumerate}

\item {\bf Claims}
\item[] Question: Do the main claims made in the abstract and introduction accurately reflect the paper's contributions and scope?
\item[] Answer: \answerYes{}
\item[] Justification: The abstract states the benchmark size, the seven open-weight models, the 1{,}680 double-annotated responses, and the 79.3--95.3\% confident-fabrication range on C1--C3; the introduction presents the same conclusion more concisely, stating that six of the seven models exceed 79\%. Section~6 marks what the results do not cover: one neutral Bengali instruction, one language, and the Hanafi position encoded in the C3 gold answers.

\item {\bf Limitations}
\item[] Question: Does the paper discuss the limitations of the work performed by the authors?
\item[] Answer: \answerYes{}
\item[] Justification: Section~6 ends with a \textbf{Limitations} paragraph covering the single neutral instruction, the Bengali-only scope, the 240-item benchmark size, and the Hanafi position adopted for the fiqh gold answers. It also states how Table~2 should be read: NA responses stay in the denominator, so a model that often ends without stating a claim records a lower fabrication rate without being any more accurate. TigerLLM-1B-it is that case, with 43 NA responses on C1--C3 and no correctly sourced answer. The same paragraph states that the benchmark measures fabrication and is not a source for religious rulings.

\item {\bf Theory assumptions and proofs}
\item[] Question: For each theoretical result, does the paper provide the full set of assumptions and a complete (and correct) proof?
\item[] Answer: \answerNA{}
\item[] Justification: The paper reports an empirical evaluation and states no theorems, lemmas, or proofs.

\item {\bf Experimental result reproducibility}
\item[] Question: Does the paper fully disclose all the information needed to reproduce the main experimental results of the paper to the extent that it affects the main claims and/or conclusions of the paper (regardless of whether the code and data are provided or not)?
\item[] Answer: \answerYes{}
\item[] Justification: Section~4.1 names the seven evaluated models. Section~4.2 gives the instruction condition, greedy deterministic decoding, per-model chat-template resolution, and which runs were local and which used a hosted API. Appendix~A reproduces the instruction as issued; Appendix~D reports the compute configuration and per-model running times. Model repository and revision identifiers, software environment details, evaluation code, and run commands are provided in the public \href{https://github.com/Meowthhh/BILAL-Bench}{GitHub} repository alongside the benchmark itself, so the pipeline can be rerun end to end.
\item[] Guidelines:
\begin{itemize}
\item The answer \answerNA{} means that the paper does not include experiments.
\item If the paper includes experiments, a \answerNo{} answer to this question will not be perceived well by the reviewers: Making the paper reproducible is important, regardless of whether the code and data are provided or not.
\item If the contribution is a dataset and\slash or model, the authors should describe the steps taken to make their results reproducible or verifiable.
\item Depending on the contribution, reproducibility can be accomplished in various ways. For example, if the contribution is a novel architecture, describing the architecture fully might suffice, or if the contribution is a specific model and empirical evaluation, it may be necessary to either make it possible for others to replicate the model with the same dataset, or provide access to the model. In general, releasing code and data is often one good way to accomplish this, but reproducibility can also be provided via detailed instructions for how to replicate the results, access to a hosted model (e.g., in the case of a large language model), releasing of a model checkpoint, or other means that are appropriate to the research performed.
\item While NeurIPS does not require releasing code, the conference does require all submissions to provide some reasonable avenue for reproducibility, which may depend on the nature of the contribution. For example:
\begin{enumerate}
\item If the contribution is primarily a new algorithm, the paper should make it clear how to reproduce that algorithm.
\item If the contribution is primarily a new model architecture, the paper should describe the architecture clearly and fully.
\item If the contribution is a new model (e.g., a large language model), then there should either be a way to access this model for reproducing the results or a way to reproduce the model (e.g., with an open-source dataset or instructions for how to construct the dataset).
\item We recognize that reproducibility may be tricky in some cases, in which case authors are welcome to describe the particular way they provide for reproducibility. In the case of closed-source models, it may be that access to the model is limited in some way (e.g., to registered users), but it should be possible for other researchers to have some path to reproducing or verifying the results.
\end{enumerate}
\end{itemize}

\item {\bf Open access to data and code}
\item[] Question: Does the paper provide open access to the data and code, with sufficient instructions to faithfully reproduce the main experimental results, as described in supplemental material?
\item[] Answer: \answerYes{}
\item[] Justification: \textsc{Bilal-Bench}, its annotation materials, evaluation and analysis code, exact run commands, and a pinned \texttt{requirements.txt} are publicly available in the \href{https://github.com/Meowthhh/BILAL-Bench}{GitHub} repository.

\item {\bf Experimental setting/details}
\item[] Question: Does the paper specify all the training and test details (e.g., data splits, hyperparameters, how they were chosen, type of optimizer) necessary to understand the results?
\item[] Answer: \answerYes{}
\item[] Justification: Nothing is trained or tuned; the seven models are evaluated as released. Section~4.2 states the decoding configuration, the absence of a system message or exemplars, per-model chat-template resolution, the split between local and hosted-API execution, the 512-token budget with a single rerun at 1{,}024 tokens for length-capped responses, and the fixed rule that trims self-generated follow-up turns. Checkpoint and revision identifiers, package versions, environment details, and run commands are documented in the public \href{https://github.com/Meowthhh/BILAL-Bench}{GitHub} repository.

\item {\bf Experiment statistical significance}
\item[] Question: Does the paper report error bars suitably and correctly defined or other appropriate information about the statistical significance of the experiments?
\item[] Answer: \answerYes{}
\item[] Justification: Table~2 reports 95\% Wilson confidence intervals for the confident-fabrication rate over all 150 C1--C3 responses per model, with nothing excluded. The intervals treat the 150 items as Bernoulli evaluation units, so they quantify item-level uncertainty in the observed proportion rather than run-to-run or decoding variance; decoding is greedy, which leaves no decoding variance to estimate. Table~3 uses the complete category-specific denominators. Label reliability is reported separately in Section~3.3 as inter-annotator agreement of Cohen's $\kappa = 0.864$ on the eight-class taxonomy and $0.893$ collapsed.

\item {\bf Experiments compute resources}
\item[] Question: For each experiment, does the paper provide sufficient information on the computer resources (type of compute workers, memory, time of execution) needed to reproduce the experiments?
\item[] Answer: \answerYes{}
\item[] Justification: Section~4.2 states that local models ran on Kaggle GPUs and that GPT-OSS-120B was accessed through a hosted API. Appendix~D gives the hardware and the timings: the Kaggle session exposed 2$\times$ NVIDIA T4 GPUs with 16~GB VRAM each, every local run used a single T4, the six local runs totalled 814.7 minutes (13.58 single-GPU hours), and the hosted run took 34.5 minutes, for 849.2 minutes (14.15 hours) of recorded wall-clock time across the seven models. Table~6 breaks that total down per model. This covers all 1{,}680 responses; no training or hyperparameter search was run.

\item {\bf Code of ethics}
\item[] Question: Does the research conducted in the paper conform, in every respect, with the NeurIPS Code of Ethics \url{https://neurips.cc/public/EthicsGuidelines}?
\item[] Answer: \answerYes{}
\item[] Justification: We have read the NeurIPS Code of Ethics, and the study conforms to it. The work measures fabrication in order to document a risk to an under-evaluated language community. Gold references were drawn from established Bengali editions and fatwa portals and checked by a qualified reviewer (Section~3.2), and the release carries our own question text and verification notes rather than redistributed editions of those works. Sections~1 and~6 state that the work neither encourages nor supports using language models to obtain religious rulings. Both annotators are members of the research team and remain anonymous, and no personal data were collected at any stage.
\item[] Guidelines:
\begin{itemize}
\item The answer \answerNA{} means that the authors have not reviewed the NeurIPS Code of Ethics.
\item If the authors answer \answerNo{}, they should explain the special circumstances that require a deviation from the Code of Ethics.
\item The authors should make sure to preserve anonymity (e.g., if there is a special consideration due to laws or regulations in their jurisdiction).
\end{itemize}

\item {\bf Broader impacts}
\item[] Question: Does the paper discuss both potential positive societal impacts and negative societal impacts of the work performed?
\item[] Answer: \answerYes{}
\item[] Justification: Section~1 states the harm the work is about: false authority placed behind text that readers treat as sacred. Sections~5 and~6 identify manufactured citations as the more dangerous failure mode, because a s\=urah:ayah number, or a collection, narrator, and hadith number, is the provenance signal a non-specialist would use to check an answer, and 1{,}316 of the 1{,}320 F1--F4 responses are assertive rather than hedged. On the other side, the release gives an evaluation instrument to a community whose exposure to this failure mode had not been measured, and it pairs questions with verified reference answers, so it supports measuring fabrication rather than producing more of it.
\item[] Guidelines:
\begin{itemize}
\item The answer \answerNA{} means that there is no societal impact of the work performed.
\item If the authors answer \answerNA{} or \answerNo{}, they should explain why their work has no societal impact or why the paper does not address societal impact.
\item Examples of negative societal impacts include potential malicious or unintended uses (e.g., disinformation, generating fake profiles, surveillance), fairness considerations (e.g., deployment of technologies that could make decisions that unfairly impact specific groups), privacy considerations, and security considerations.
\item The conference expects that many papers will be foundational research and not tied to particular applications, let alone deployments. However, if there is a direct path to any negative applications, the authors should point it out. For example, it is legitimate to point out that an improvement in the quality of generative models could be used to generate Deepfakes for disinformation. On the other hand, it is not needed to point out that a generic algorithm for optimizing neural networks could enable people to train models that generate Deepfakes faster.
\item The authors should consider possible harms that could arise when the technology is being used as intended and functioning correctly, harms that could arise when the technology is being used as intended but gives incorrect results, and harms following from (intentional or unintentional) misuse of the technology.
\item If there are negative societal impacts, the authors could also discuss possible mitigation strategies (e.g., gated release of models, providing defenses in addition to attacks, mechanisms for monitoring misuse, mechanisms to monitor how a system learns from feedback over time, improving the efficiency and accessibility of ML).
\end{itemize}

\item {\bf Safeguards}
\item[] Question: Does the paper describe safeguards that have been put in place for responsible release of data or models that have a high risk for misuse (e.g., pre-trained language models, image generators, or scraped datasets)?
\item[] Answer: \answerNA{}
\item[] Justification: The released artefact is an evaluation benchmark: questions paired with expert-verified reference answers. It ships no model weights and presents no generated model output as authoritative religious guidance, so the controlled-release safeguards this question contemplates do not apply.

\item {\bf Licenses for existing assets}
\item[] Question: Are the creators or original owners of assets (e.g., code, data, models), used in the paper, properly credited and are the license and terms of use explicitly mentioned and properly respected?
\item[] Answer: \answerYes{}
\item[] Justification: The manuscript credits the model families it evaluates and, in Section~3.2, the reference works consulted for the gold answers. The public \href{https://github.com/Meowthhh/BILAL-Bench}{GitHub} repository lists the revision, source URL, licence name, and applicable terms of use for each evaluated model. The released benchmark carries our own Bengali question text and verification notes rather than redistributed editions of the consulted works.

\item {\bf New assets}
\item[] Question: Are new assets introduced in the paper well documented and is the documentation provided alongside the assets?
\item[] Answer: \answerYes{}
\item[] Justification: \textsc{Bilal-Bench} is documented in the paper itself: Table~1 gives the category composition and the behaviour expected in each category, Section~3.2 the curation and expert-verification procedure, Section~3.3 the two-axis rubric and the blind double-annotation protocol, and Appendices~A--B the instruction as issued and one released item per category. The \href{https://github.com/Meowthhh/BILAL-Bench}{GitHub} repository README documents the dataset, together with the per-item verification notes and the licence.

\item {\bf Crowdsourcing and research with human subjects}
\item[] Question: For crowdsourcing experiments and research with human subjects, does the paper include the full text of instructions given to participants and screenshots, if applicable, as well as details about compensation (if any)?
\item[] Answer: \answerNA{}
\item[] Justification: Two members of the research team labelled all 1{,}680 responses independently and blind to model identity, under a written rubric whose boundary rules were fixed before annotation began. No crowdsourcing platform and no external participants were involved, and no personal data were collected, so the question does not apply.

\item {\bf Institutional review board (IRB) approvals or equivalent for research with human subjects}
\item[] Question: Does the paper describe potential risks incurred by study participants, whether such risks were disclosed to the subjects, and whether Institutional Review Board (IRB) approvals (or an equivalent approval/review based on the requirements of your country or institution) were obtained?
\item[] Answer: \answerNA{}
\item[] Justification: Nobody was recruited as a study participant. Two research-team members labelled model outputs internally, with no intervention and no collection of personal data, so there were no human subjects for whom IRB approval or risk disclosure would apply.

\item {\bf Declaration of LLM usage}
\item[] Question: Does the paper describe the usage of LLMs if it is an important, original, or non-standard component of the core methods in this research? Note that if the LLM is used only for writing, editing, or formatting purposes and does \emph{not} impact the core methodology, scientific rigor, or originality of the research, declaration is not required.
\item[] Answer: \answerNA{}
\item[] Justification: Language models are the object of study here, not a component of the method; Sections~4 and~5 describe how they were evaluated. No language model wrote benchmark items, produced or adjudicated labels, designed the experiments, or computed any reported number; all labels, analyses, and conclusions are the authors' own. ChatGPT was used only to polish the manuscript for clarity, grammar, and style. 

\end{enumerate}

\end{document}